\documentclass[runningheads]{llncs}
\usepackage{graphicx}
\usepackage{array}
\usepackage{multirow}
\usepackage{adjustbox}
\usepackage{float}
\usepackage{marvosym}

\begin{document}
\title{Actively evaluating and learning the distinctions that matter: Vaccine safety signal detection from emergency triage notes}
\titlerunning{Vaccine Safety Signal Detection from Triage Notes}
%
\author{Sedigh Khademi\inst{1,2,4}\textsuperscript{\Letter}\orcidID{0000-0001-6146-1415} \and
Christopher Palmer\inst{1,2}\orcidID{0000-0001-6554-9027} \and
Muhammad Javed\inst{1,2}\orcidID{0000-0002-7022-6596} \and\
Hazel J Clothier\inst{1,2,4,5}\orcidID{0000-0001-7594-0361}\and
Jim P Buttery\inst{1,2,3,4}\orcidID{0000-0001-9905-2035}\and
Gerardo Luis Dimaguila\inst{1,2,4}\orcidID{0000-0002-3498-6256}\and\
Jim Black\inst{6}\orcidID{0000-0002-9287-8712}}
\authorrunning{S. Khademi et al.}
%
\institute{Centre for Health Analytics, Murdoch Children's Research Institute, Parkville, Australia
\\
{\{sedigh.khademi, chris.palmer, muhammad.javed, hazel.clothier,
jim.buttery, gerardoluis.dimaguil\}@mcri.edu.au}\\
\url{} \and
Epidemiology Informatics Group and Surveillance of Adverse Events Following Vaccination in the Community, MCRI, Parkville, Australia\\
\email
\and
Infectious Disease Unit, Department of General Medicine, Royal Children’s Hospital, Parkville, Australia\\
\and
Department of Paediatrics, The University of Melbourne, Parkville, Australia\\
\and
Melbourne School of Population and Global Health, University of Melbourne, Melbourne, Australia \\
\email jim.black@health.vic.gov.au \\
\and
Department of Health, State Government of Victoria, Melbourne, Australia\\ }
\maketitle              
\begin{abstract}
The rapid development of COVID-19 vaccines has showcased the global community's ability to combat infectious diseases. However, the need for post-licensure surveillance systems has grown due to the limited window for safety data collection in clinical trials and early widespread implementation. This study aims to employ Natural Language Processing (NLP) techniques and Active Learning (AL) to rapidly develop a classifier that detects potential vaccine safety issues from emergency department (ED) notes. ED triage notes, containing expert, succinct vital patient information at the point of entry to health systems, can significantly contribute to timely vaccine safety signal surveillance. While keyword-based classification can be effective, it may yield false positives and demand extensive keyword modifications. This is exacerbated by the infrequency of vaccination-related ED presentations and their similarity to other reasons for ED visits. NLP offers a more accurate and efficient alternative, albeit requiring annotated data, which is often scarce in the medical field. Active learning optimizes the annotation process and the quality of annotated data, which can result in faster model implementation and improved model performance. This work combines active learning, data augmentation, and active learning and evaluation techniques to create a classifier that is used to enhance vaccine safety surveillance from ED triage notes.

\keywords{Natural language processing  \and Active learning  \and Large language models  \and Emergency department  \and Vaccine safety  \and Post-licensure surveillance .}
\end{abstract}
\section{Introduction}
\subsection{Background}
The fast-paced development and adoption of vaccines in recent years has highlighted the ability of the international community to respond to global infectious disease threats. This included regulatory provisions employing emergency use authorizations or provisional registrations for vaccines following completion of clinical trials. However, the narrow window for collecting safety data in clinical trials data used for these authorizations amplified the critical role of post-licensure surveillance systems. An Adverse Event Following Immunization (AEFI) is any untoward medical occurrence that follows vaccination \cite{council2012application}. AEFI can be mild, such as pain, redness, or swelling at the injection site; more serious, such as allergic reactions or seizures; and can occur up to several weeks following vaccination, for example with neurological conditions like Guillain-Barré syndrome \cite{choe2011serious}. Therefore, monitoring vaccine safety from as many sources as possible ensures timely detection of any adverse events or unexpected patterns \cite{buttery2022information,crawford2014active}. Emergency department (ED) presentations are an important source of data for vaccine safety surveillance because they document one of the initial points of contact with the healthcare system for individuals.

Natural Language Processing (NLP) techniques harness the power of computing systems that can learn the structure of language to become capable of identifying specifics in health-related discussions \cite {botsis2023improving}. However, these systems typically require lots of text examples to learn from, and assembling clinical-based examples can be difficult due to the time and cost required for the clinical expertise needed to identify examples. An approach called Active Learning (AL) \cite {settles2012active} can guide the identification of optimum examples so that the burden of labeling is minimized. In this work we use AL and other techniques to develop data and train an operational NLP model that is performing real-time identification of potential vaccine safety issues in emergency department notes. 

\subsection{Related work}
Emergency department triage notes consist of brief phrases created during the initial assessment of a patient presenting to a hospital. They typically include the patient's medical history, presenting symptoms, and the chief complaint. This information is used to prioritize clinical attention to the patient and can also be used to track trends in patient visits and to improve the quality of care. Research has shown that triage notes, whether used  in isolation or combined with diagnostic codes, have the potential to significantly enhance the  identification of health information, which may not always be represented in such detail in the codified information in electronic medical records \cite{arnaud2021nlp}.

Several approaches have been employed for classification of triage notes to detect cases, including keyword-based, linguistics-driven, statistical, and machine learning algorithms, as well as combinations of these \cite{conway2013using} . Keyword-based classification, or pattern-matching, is a simple and effective method for classifying presentations of exceptionally rare health conditions. However, for more common health conditions, it can lead to false positives, as the keywords used for classification may also appear in notes for other conditions. Additionally, the keywords need to be modified extensively to address misspellings and variances in triage terminology, which can be time-consuming \cite{conway2013using}. Natural language processing (NLP) provides a promising option for more accurate and efficient medical text classification and has been used in this domain in recent years \cite{picard2022emergency}. The implementation of NLP techniques requires access to annotated data, which can be challenging to acquire in the medical domain due to the scarcity of resources, including expert domain knowledge and time. Researchers in the application of machine learning-based syndromic surveillance of ED notes have had to manually label thousands of records \cite{rozova2022detection} and have observed that difficulties in obtaining data and domain expertise across varied institutions has affected the generalizability of the models they have developed \cite{dexter2020generalization,spasic2020clinical}. Furthermore, training data obtained through pattern matching and sampling has not been sufficiently informative to train robust classifiers. Data augmentation strategies have been required to introduce variety and expand the training data to achieve satisfactory results \cite{khademi2023data}.

Active learning (AL) is an approach that leverages machine learning models to intelligently select and prioritize which data points should be annotated by a human “Oracle” who is normally a domain expert. By doing so, the use of AL not only maximizes the efficiency of the annotation process but also assists to produce training data that enhances the overall performance of models \cite{Olsson1042586}. Active learning is a well-established concept  in machine learning \cite{settles2012active} and has found applications in many clinical NLP tasks including text classification \cite{figueroa2012active,mottaghi2020medical}. The pool-based method represents the most prevalent active learning approach. In this method, a pool of unlabeled data is initially available, and the algorithm uses a query strategy to iteratively select data points from the pool for annotation. There are two main approaches for query strategies. The first is informativeness-based, where an informative measure is assigned to each data point and the instances with the highest measure are subsequently chosen. Uncertainty sampling \cite{lewis1995sequential} is a commonly used strategy in this category \cite{zhang2022survey}.  The second query category uses representativeness-based approaches, or diversity sampling, which focuses on selecting a diverse set of data points that represent the underlying distribution of the entire data set.

Despite its advantages, the effectiveness of active learning can be challenged under certain conditions. For instance, AL-based classifiers' performances could be limited with increased class complexity or when the data set is imbalanced \cite{attenberg2013class}. To address these limitations, data augmentation strategies can be employed. These strategies help improve a classifier’s performance by generating additional data that ensures a more balanced and comprehensive data set \cite{fonseca2023improving,moles2024exploring}. Studies have shown that when limited labeled data are available, creating label-flipped data is more effective for data augmentation than creating label-preserved data \cite{zhou2021flipda}. This could be done by tasking humans with revising documents to align with counterfactual labels while preserving coherence with minimal changes \cite{kaushik2019}.

Beyond training data, careful consideration must also be given to the evaluation of machine learning models. When annotation is expensive, the principle of carefully selecting data points for labeling extends to test data as well \cite{kossen2021active}. Actively choosing test examples for labeling ensures that resources are used effectively, and that model evaluation is both accurate and valuable. Moreover, developing reliable operational machine learning models requires robust evaluation to ensure production readiness \cite{breck2017ml}. Ideally, this evaluation should be conducted using data from the deployment environment \cite{ha2021alt}.

To address both the labeled data scarcity issue and the requirement for data representativeness, a promising strategy is the utilization of active learning methodologies \cite{zhang2022survey}. This paper outlines a process for rapidly developing a syndromic surveillance classifier using data from various ED departments. By combining active learning, data augmentation, and guided evaluation, we aim to minimize the need for extensive domain expert involvement.
\subsection{Objectives}
Potential adverse events following immunization (AEFI) presentations are a very small proportion of cases in emergency departments and developing an initial labeled data set for training a classifier to detect these cases is a challenging task. In this work, we efficiently select the most valuable data for fine-tuning a high-performing pre-trained language model for detecting potential AEFI in emergency department notes by:

•	Selecting initial training data using a topic modelling technique with sentence vectors.

•	Using uncertainty-based active learning strategies to select the data points to be labeled by the Oracle.

•	Augmenting the training data using counterfactually labeled data.

•	Actively evaluating the model against the deployment environment data. 
\section{Data}
The data comes from the SynSurv syndromic surveillance system, which continuously receives data from a majority of the public hospitals with Emergency Departments in Victoria, Australia. These data include texts written by ED nurses at the triage desk after their initial assessment of each patient. The texts are typically short and use abbreviations and phrases that are specific to the medical field. They vary in length and quality but typically include the patient's complaint, medical history, relevant negatives, and nurse's observations. The data also include datetime, the patient’s age and sex,
and hospital identification.

The unlabeled text pool consisted of 35,998 ED unique triage texts at least 3 characters in length, that were received by SynSurv from 1 January 2021 to 13 June 2023, and filtered on vaccine related terms. These included ‘vacc’ and ‘vax’, a list of vaccination brands used in Australia, and a list of diseases for which the population receives vaccinations \cite{AustralianImmunisationHandbook2024}. From this we removed records that were included just because of COVID-19 vaccination history with mentions like ‘fully vaxed’, ‘triple vaxed’ or ‘covid vaccinated’. This reduced the number of records to 11,060 that were potentially AEFI-focused. We additionally sampled 61,600 recent records from the SynSurv system, which we refer to as the deployment-environment text pool and these were used in later rounds of model development to help to evaluate and guide model development. Table 1 presents examples of AEFI and non-AEFI related presentations.
\vspace{10pt} 
\begin{table}
\centering
\caption{Examples of AEFI in ED triage notes}
\begin{tabular}{|l|c|}
\hline
\textbf {Presentation (not actual text but illustrative examples)    }             & \textbf {AEFI} \\ \hline
onset chest pain with palpitations. had 2nd dose of pfizer yesterday.    & Yes  \\
central dull ache pain since. rr 20 sats 98 hr135                        &      \\ \hline
red blanching rash body, started yesterday. rash worse today.            & Yes  \\
not itchy. chicken pox vaccine 1/52. fever 4/7 ago. dec solids intake.   &      \\ \hline
dizzy, giddy, lightheaded, vomiting, impaired gait at 0000. covid vax x2 & No   \\ \hline
monkey bite l) hand 1/52. had 1st dose of rabies vaccine 5/7.            & No   \\
was asked to represent today for 2nd dose                                &      \\ \hline
\end{tabular}
\end{table}

\subsection{Data pre-processing}
Since the data come from many hospitals, there is significant variability in the text. Some texts contain repetitive diagnostic or administrative fragments and embedded tabular information, both of which needed removal. Each triage note was prefixed with the patient's age and sex since vaccine adverse events can vary based on these factors. For example, heart-related issues associated with the mRNA COVID-19 vaccines are more common in young males \cite{alami2022risk}.

\subsection{Labeling guidelines}
A record is labeled as AEFI-related if it contains mentions of vaccine-related keywords in the presentation, and the patient or a trained healthcare worker (including a doctor, nurse, or paramedic), has made a connection between the patient's condition and a recently received vaccine. A record is not labeled as AEFI-related if the vaccination mention is only a report on vaccination status or of the need to receive a vaccine. Nor is it labeled as an AEFI if the vaccination mention pertains solely to past medical history – there needs to be a mention of recent vaccination. However, a record will be labeled as AEFI-related if the historical vaccination mention is specifically linked in the text as related to the presenting problem.
Our data set included the COVID-19 pandemic period, a time marked by the introduction of new vaccines, significant announcements regarding COVID-19-related AEFIs  \cite{greinacher2021thrombotic,kim2021patients,schultz2021thrombosis}, the implementation of vaccine mandates, and adjustments in hospitals' policies regarding the documentation of patients' vaccination status. All these factors collectively influenced emergency presentations and triage notes, posing numerous challenges in detecting AEFI-related cases. We observed that mentions of vaccinations could appear in ED triage notes irrespective of the primary reason for the visit.

\section{Method}
\subsection{Topic modelling}
Topic modelling using BERTopic was used to select the initial candidate records for labelling from the AEFI-focused text pool. Each ED note was treated as a single unit due to its brevity and specific descriptions. For the sentence model, we created Sentence-BERT embeddings \cite{reimers2019sentence}  using the PubMedBert-base-embeddings model, which maps sentences to a 768-dimensional \cite{reimers2019sentence} vector space and is suitable for medical text. These are numerical representations of words where semantically similar words are close together in the vector space \cite{mikolov2013efficient}. We used the NLTK corpus stop words list and 2-grams with the Scikit-learn count vectorizer. We used Euclidean distance for the UMAP and HBDSCAN parameters. 

The resulting topic model gave us vector embeddings and topic clustering data, with topic centroids identified by the HBDSCAN weighted cluster centroid for each topic. The model identified a total of 30 topics, but we reduced this to 29 via the BERTopic function to reduce outliers. These topic clusters with their respective centroids and embeddings served as the foundation for active-learning record identification processes.
\subsection{Assembling the initial data set}
To train a large language model-based classifier, we needed a training data set that was large enough to be effective, but not so large that it would be burdensome to annotate. We settled on 700 records, to allow for collecting sufficient positive examples, and which the clinician endorsed as a suitable size to label in around two hours. We identified training samples using a diversity sampling query strategy – based on selecting records along a range of their distance from the cluster centers. 
We identified 9 topics that mostly contained AEFI and found that AEFI rarely appeared in the remaining 20 topics. We sampled 60\% of the 700 records from the 9 AEFI topics and 40\% from the rest. To do this, we ordered records based on distance from the topics’ cluster centroids and sampled along that range by an interval determined by how many samples we wanted to take from each cluster. This interval depended on whether the topic was an AEFI topic, and in proportion to the cluster size, with a sampling limit to prevent exceeding the target of 700 records. Specifically, we sampled AEFI topics in proportion to the cluster size, within limits, but we sampled only 3 records from each non-AEFI topic. We considered accounting for text length variations but found it unnecessary, as most texts ranged from 30 to 50 words and topic texts were similarly sized. The data were labeled by two authors, and any disagreements were resolved through consultation with the clinician author. Table 2 presents the initial data set distribution.
\vspace{10pt} 
\begin{table}
\centering
\caption{The initial data set distribution.}
\label{tab2}
\begin{tabular}{|l|c|c|c|}
\hline
\textbf {Data set}                 & \textbf {AEFI}     & \textbf {Non-AEFI} & \textbf {Total}            \\ \hline
Training   & 266 & 296 & 562 \\ \hline
Validation   & 66 & 72 & 138 \\ \hline
\end{tabular}
\end{table}
\subsection{Model development}
We used the RoBERTa-large-PM-M3-Voc \cite{lewis2020pretrained}, a language model pre-trained on biomedical and clinical texts. This model was selected because of its outstanding performance in classifying such texts, and which we have utilized in other projects \cite {khademiJmirAI}. A model was fine-tuned from scratch for each evaluation of a data set, except for the final round where we compared a further fine-tuned model with a model trained from scratch. Models were trained for 9 epochs, with a batch size of 16, on a 32GB GPU, using defaults for the parameters. Checkpoints were saved every 10 steps, and identification of the best checkpoints was based on model loss, AUC-ROC, evaluation data F1-score, and examination of the false positives and negatives. The validation scores were used as a basis for picking the best checkpoints for evaluation. The checkpoint evaluation process took the top few checkpoints in each round, and used them to predict on the unlabeled AEFI-focused and deployment-environment text pools. Their positive AEFI predictions were then labeled. This confirmed positive predictions and identified false positives, and so indicated the direction that the data was taking the model. After determining the best checkpoints by comparing their performance on these newly labeled records, the top two checkpoints’ uncertain negative predictions were also labeled – that is, where their probabilities were less than 90\%. This is the same approach used by the AL uncertainty sampling query strategy, which based on our prior experience is the most informative strategy to assist the model to adjust its decision boundaries 
 \cite{khademi2023detecting}.  Finally, any potential false negative predictions were identified by looking for AEFI-related words or phrases in the high-probability \begin{math}(\geq 90\%)\end{math} negative predictions of the checkpoints. These labeling steps not only allowed for model evaluation but also identified candidate records for adding to training data.

In each phase, we took informative examples to add into our training data, being the false positive predictions and the uncertain negatives. We also added enough correctly predicted (and labeled) positive examples from the text pools to ensure that there was no more than a 3:2 ratio between negative and positive examples. For rounds 2 and 3 we added some of the newly labeled examples into the validation data, but we did not add any to validation for the 4th round.

\subsection{Data augmentation}
We created additional data points by using label flipping augmentation technique. This consisted of the addition or removal of AEFI-specific text to create a synthetic record that was very like the record it was derived from, with the difference being in the AEFI-specific information. These pairs were then labeled as positive and negative examples.
Going into the second round of training (i.e., after assessing the initial model), we saw that incorrect predictions were almost always false positives – so we included 100 augmented synthetic negative examples that were derived from sampled correct positive records, and then removing the text that contained the AEFI information. For example, to classify a description of abdominal pain as an AEFI, a mention of the flu vaccine is necessary. In the text, “abdominal pain for 1/52 - onset 30 minutes post flu vaccine. 1 x vomit and 2 x loose stools,” the vaccine mention indicates a potential connection to the symptoms.To help the model understand the significance of the vaccine reference, a synthetic negative example was created by removing it, to say “abdominal pain for 1/52 - onset 30 minutes. 1 x vomit and 2 x loose stools.”

In the third training round, when assessing the new model’s checkpoints on deployment-environment text pool data, we again observed a tendency to predict false positives. However, we did not use any true-positive records; instead, we used the false-positive examples and added AEFI-related text to create a synthetic label-flipped true-positive equivalent. For example, “flu sx for 1/52, today pain when coughing” was used to derive a correct AEFI example by adding information about a flu vaccine: “flu sx for 1/52, had flu vaccine 1/52. today pain when coughing”. The model then had an opportunity to see that a description of flu is not enough to make an AEFI prediction – a flu vaccine mention is required. We added 83 new synthetic records, with 67 of these derived from records from the deployment-environment text pool, the rest from the AEFI-focused text pool.

Evaluation of the results from the third round showed us that the model had erroneously learned that certain phrases were indicative of AEFI and so it was predicting false positives when finding them. To fix this for the fourth round of training we added more examples from the deployment-environment text pool that contained these phrases and removed 6 synthetic records in the training data that contained these as a positive example. Table 3 shows the evolution of the training and validation data over the four rounds of training, the column “Syn” describes the percentage of synthetic records contained in each data set.
\vspace{10pt} 
\begin{table}
\centering
\caption{The development of training data over 4 rounds}
\begin{adjustbox}{width=\textwidth}
\begin{tabular}{>{\centering\arraybackslash}p{0.11\linewidth}|>{\centering\arraybackslash}p{0.11\linewidth}>{\centering\arraybackslash}p{0.11\linewidth}>{\centering\arraybackslash}p{0.11\linewidth}>{\centering\arraybackslash}p{0.11\linewidth}|>{\centering\arraybackslash}p{0.11\linewidth}>{\centering\arraybackslash}p{0.11\linewidth}>{\centering\arraybackslash}p{0.11\linewidth}>{\centering\arraybackslash}p{0.11\linewidth}|}
\cline{2-9}
                              & &&\textbf{Training}&                                                           & &&\textbf{Validation}&                                                        \\ \cline{2-9} 
                              & \multicolumn{1}{l|}{Pos} & \multicolumn{1}{l|}{Neg} & \multicolumn{1}{l|}{\textbf{Total}} & Syn  & \multicolumn{1}{l|}{Pos} & \multicolumn{1}{l|}{Neg} & \multicolumn{1}{l|}{\textbf{Total}} & Syn \\ \hline
\multicolumn{1}{|l|}{Round 4} & \multicolumn{1}{l|}{630} & \multicolumn{1}{l|}{827} & \multicolumn{1}{l|}{\textbf{1457}}  & 12\% & \multicolumn{1}{l|}{102} & \multicolumn{1}{l|}{122} & \multicolumn{1}{l|}{\textbf{224}}   & 8\% \\ \hline
\multicolumn{1}{|l|}{Round 3} & \multicolumn{1}{l|}{604} & \multicolumn{1}{l|}{787} & \multicolumn{1}{l|}{\textbf{1391}}  & 13\% & \multicolumn{1}{l|}{102} & \multicolumn{1}{l|}{122} & \multicolumn{1}{l|}{\textbf{224}}   & 8\% \\ \hline
\multicolumn{1}{|l|}{Round 2} & \multicolumn{1}{l|}{456} & \multicolumn{1}{l|}{551} & \multicolumn{1}{l|}{\textbf{1007}}  & 9\%  & \multicolumn{1}{l|}{93}  & \multicolumn{1}{l|}{113} & \multicolumn{1}{l|}{\textbf{206}}   & 4\% \\ \hline
\multicolumn{1}{|l|}{Round 1} & \multicolumn{1}{l|}{266} & \multicolumn{1}{l|}{296} & \multicolumn{1}{l|}{\textbf{562}}   &      & \multicolumn{1}{l|}{66}  & \multicolumn{1}{l|}{72}  & \multicolumn{1}{l|}{\textbf{138}}   &     \\ \hline
\end{tabular}
\end{adjustbox}
\end{table}
\vspace{10pt} 
\subsection{Model evaluation}
As training progressed into rounds 3 and 4, we labeled and added examples to the training data from both text pools but relied mostly on the deployment-environment text pool for evaluation as it was representative of the balance of data the eventual model would encounter. Although there was mostly a consensus between the models’ checkpoints on this data when correctly identifying AEFI, there were more opportunities for differences in the false positives due to the variety of negative examples. As a result, more records needed to be labeled to encompass the predictions of the top few checkpoints per round. We had to label around 1400 records from the deployment-environment text pool data over rounds 3 and 4 of training to be able to compare our models, with the benefit of obtaining more potential labeled training data.

For each round we trained new models from scratch on the expanded training data set of each round. However, as models were scoring at an F1-score of nearly 0.9 by the end of round 3, for round 4 we also assessed further fine-tuning a model loaded from the best round 3 checkpoint. We found that despite the round 4 model that was trained from scratch performing better during training, on the deployment-environment text pool the further fine-tuned round 3 model out-performed the checkpoints from the round 4 model fine-tuned from scratch. The further fine-tuned model achieved an F1-score of 0.97 compared to the 0.90 from the best checkpoint of the from-scratch model.

\section{Results}
Validation data was used for initial assessment of the best checkpoints. Loss, AUC-ROC, and F1-scores were considered vs. the validation data, which led us to consider several checkpoints from each training round for further evaluation. The best validation F1-scores were between 0.92 and 0.94 in each training round, which included extra validation records in each round, until round 4. 
We then tested the best checkpoints against both the AEFI-focused and the deployment-environment text pools. The best assessment of progress came from the deployment-environment data, as using that data allowed us to guide model development into the domain it would typically encounter. The AEFI-focused data became increasingly less helpful because we were training the models about the larger domain, and we were also removing newly labeled AEFI-focused records from the text pool to add to our training data. Even with additional labeling, the available labeled records in the AEFI-focused text pool shrank from 893 to 115 over the course of the four training rounds as newly labeled records were moved into the training data.

After eventually labeling all positive predictions and uncertain negatives from the best models from round 2 onwards in the deployment-environment data, we ended up with 1,806 records to use for evaluation – 107 positive and 1699 negative. During dataset finalization, we used pattern matching to identify misclassified AEFI records and found very few, so we gained confidence that the best models captured most true positives. 

When comparing all models to the final labeled cohort, we saw significant score improvements from 0.82 to 0.97, mainly due to precision rising from 0.72 to 0.96. The round 3 model, further fine-tuned for the round 4 model, gained its F1-score improvement primarily from an increase in precision. We also included the scores of pattern-matching as a baseline comparison. These results are shown in Table 4, where the round 1 model is included where its predictions overlapped the round 2 models. F1Beta in the table uses a beta of 1.3 to slightly favor recall.
\vspace{10pt} 
\begin{table}[H]
\centering
\caption{Model scores on the deployment-environment text pool}
\begin{tabular}{|>{\centering\arraybackslash}p{0.15\linewidth}|>{\centering\arraybackslash}
p{0.09\linewidth}|>{\centering\arraybackslash}p{0.09\linewidth}|>{\centering\arraybackslash}p{0.09\linewidth}|>{\centering\arraybackslash}p{0.09\linewidth}|>{\centering\arraybackslash}p{0.12\linewidth}>{\raggedright\arraybackslash}p{0.15\linewidth}|>{\centering\arraybackslash}p{0.1\linewidth}|>{\centering\arraybackslash}p{0.08\linewidth}|>{\centering\arraybackslash}p{0.1\linewidth}|}
\hline
\textbf{Model} & \textbf{TP} & \textbf{TN} & \textbf{FN} & \textbf{FP} & \multicolumn{2}{c|}{\textbf{Precision}} & \textbf{Recall} & \textbf{F1} & \textbf{F1Beta} \\ \hline
Round 4        & 106         & 1694        & 1           & 5           & \multicolumn{2}{c|}{0.955}              & 0.991           & 0.972       & 0.977           \\ \hline
Round 3        & 106         & 1672        & 1           & 27          & \multicolumn{2}{c|}{0.787}              & 0.991           & 0.883       & 0.909           \\ \hline
Round 2        & 101         & 1660        & 6           & 39          & \multicolumn{2}{c|}{0.721}              & 0.944           & 0.818       & 0.847           \\ \hline
Round 1        & 101         & 1223        & 6           & 476         & \multicolumn{2}{c|}{0.175}              & 0.944           & 0.295       & 0.359           \\ \hline
Pattern Matching            
   & 85         & 1662        & 22           & 37         & \multicolumn{2}{c|}{0.697}              & 0.794           & 0.742       & 0.755           \\ \hline
\end{tabular}
\end{table}
\vspace{10pt} 

\section{Discussions}
\subsection{Overview}
The success of our approach was that we used active learning to cold-start a classification process that led to the development of initial models that readily identified AEFI, then combined active learning and human involvement to develop training data and evaluate models that progressively learned the target deployment environment. Although the early models were imprecise, the human-centered oversight of model development using effective AL-based sampling strategies, data augmentation, and a guided evaluation approach resulted in a very precise model that was robust enough to deploy and is proving itself daily.  Our work demonstrated the utility and efficiency of an application-centric evaluation approach in refining our model for AEFI detection. Application-centric evaluations focus on how a model operates within its intended use case, emphasizing real-world applicability over theoretical performance metrics \cite{hutchinson2022evaluation}.
\subsection{Findings}
Our work confirmed that an AL sentence embedding clustering technique is an effective approach to create a balanced representative data set from a label-centric text pool for the initial training of a classifier. The classifier initially achieved a high recall on the target AEFI label validation data but with low precision. This aligns with findings from other studies, which have shown that clustering unlabeled data and selecting data points from each cluster enables the creation of an initial data set that can rapidly achieve high performance with fewer additional queries \cite{kang2004using}.

Use of the initial classifier’s predictions enabled identification of further training records using an AL uncertainty-sampling combined with a human-guided data augmentation (label flipping) approach. We saw the evolution of increasingly precise models with high recall over several rounds of training using this labeling approach on both label-centric data and increasingly on data obtained from the deployment environment.  

Human-in-the-loop model development which involves a collaborative process where human feedback enhances the performance and applicability of ML solutions, is crucial for effective outcomes \cite{andersen2023we,kaushik2019,kottke2018other}. Our research demonstrated that involving humans not only in labeling data but also in actively intervening with the data to align records with different labels helps uncover the critical distinctions that contribute to developing a robust classifier. Training classifiers that don’t make spurious associations is particularly important when working with highly specialized medical domain texts, where symptoms, diagnoses, and past medical histories can be similar among many documents.

A crucial factor in our success was tackling the challenge of sample-efficient learning, where no ground truth test set existed and high-quality labeled data for evaluation was limited, by using deployment environment data to perform evaluation of our AL based model. We predicted on the data using our best models and took their positive predictions and uncertain negative predictions, supported by pattern matching to find possible missed examples. These records were labeled to provide a growing pool of evaluation data and potential extra training data until we reached the point in our training cycles where almost all positive labels were being successfully identified by the models, and where minimal negative labels were being falsely identified. This approach is described in the literature as active testing or active evaluation \cite {yu2024actively}, where a ground truth does not exist, and where the performance of the active learner is evaluated by utilizing previously unlabeled data \cite{kottke2017challenges}.

We conducted error analysis on the predictions made by the best model for one month of 145,852 SynSurv records. The model predicted 286 records as AEFI-related, labeling of these showed there were 266 AEFI-records – therefore there were 20 instances of false positive predictions, around 8\% of the true positives. The primary reasons for these errors included: other uses of vaccine-related keywords, such as “had tequila shot, ? aspirated” or “had mounjaro injection on monday feeling unwell, phx: anxiety”; uses of words like vacuuming with a language similar to AEFI related presentations, for example: “exac of chronic lower back pain bkg vaccuming”;  and mentions of accidents with animal vaccinations: “self-injected with campyvax yesterday”. To check for false negatives, we reviewed records selected based on a pattern matching on vaccine-related keywords and did not find any. An analysis of the errors suggests that they stem from the minimal amount of training data that was used. For instance, there were not enough examples in the training data of animal-related vaccination records. Similarly, although there were examples of vacuuming-related negative labels, the model is still seeing the use of “vac” and making the association with vaccination when the surrounding language indicates an onset of pain, which is being interpreted as a possible AEFI. We are collecting false positives to enhance the training data and fine-tune the classifier.

Our study comes with limitations. Initially, we constructed the data pool using vaccine-related keywords, which could potentially lead to the omission of some presentations. Nevertheless, our keyword-based data selection accurately captured the space. In future work we intend to conduct topic modelling on all presentations to enhance data inclusiveness. Furthermore, in this study, for an AEFI to be predicted we require a description of a vaccination connection within the text – there is no attempt to detect AEFI using only symptoms. Our long-term strategy is to track major known AEFI reactions within emergency department notes, by developing models that can discern potential AEFI irrespective of vaccination mentions. By comparing these occurrences with baseline rates, we intend to monitor for a significant increase in such presentations. A reactions-based classifier would need to be deployed alongside a vaccination aware classifier, as a link to vaccination is required to consider a health condition a potential AEFI.

The data we obtained from various hospitals’ ED departments in our region required pre-processing to remove certain features, and furthermore, meant that models needed to learn the varied language used across these departments. Despite these challenges, the deployed model is very performant – but is highly attuned to the language of these ED departments, making it to be unlikely to be generalizable beyond our jurisdiction. Nevertheless, the AL and active evaluation approach we took is generalizable and would be effective for similar situations with initially unlabeled data containing a concentration of relevant texts. 
\section{Conclusions}
Our study demonstrated the effectiveness of integrating active learning and human oversight to develop a robust classifier. This approach, which combined initial model training with continuous human-guided data augmentation, not only improved the performance of AEFI detection but also ensured that the classifier remained adaptable and effective in a real-world application. This research underscored the value of sample-efficient learning and human-in-the-loop strategies in enhancing the safety monitoring of vaccines, ultimately contributing to better public health outcomes.
The accelerated detection of AEFI and quicker public health interventions enabled by this research contribute to strengthening vaccine monitoring systems, thereby improving public trust and confidence in vaccines.

%
%
%
\bibliographystyle{splncs04}
\bibliography{AusDM24_paper57}
%




\end{document}